\documentclass{article} 
\usepackage{iclr2025_conference,times}

\usepackage{amsmath,amsfonts,bm}

\def\eqref#1{equation~\ref{#1}}

\def\1{\bm{1}}

\DeclareMathAlphabet{\mathsfit}{\encodingdefault}{\sfdefault}{m}{sl}
\SetMathAlphabet{\mathsfit}{bold}{\encodingdefault}{\sfdefault}{bx}{n}

\usepackage{hyperref}
\usepackage{url}
\usepackage{graphicx}
\usepackage{subfig}

\title{PyViT-FUSE: A Foundation Model for Multi-Sensor Earth Observation Data}

\author{Manuel Weber\thanks{Corresponding author (manuel.wbr@gmail.com)} , Carly Beneke \\
EarthDaily Analytics \\
1055 Canada Pl \#33, \\
Vancouver, BC V6C 3L5, Canada \\
\texttt{\{manuel.weber,carly.beneke\}@earthdaily.com}
}

\iclrfinalcopy 
\begin{document}

\maketitle

\begin{abstract}

We propose PyViT-FUSE, a foundation model for earth observation data explicitly designed to handle multi-modal imagery by learning to fuse an arbitrary number of mixed-resolution input bands into a single representation through an attention mechanism. The learned patch tokens are further processed by a stack of vision transformers with a novel pyramidal structure. We train the model on a globally sampled dataset in a self-supervised manner, leveraging core concepts of the SwAV algorithm. We show the interpretability of the fusion mechanism by visualization of the attention scores and the models applicability to downstream tasks.
\end{abstract}

\section{Introduction}
\label{sec:introduction}

Foundation models (FM) for earth observations (EO) have gained traction following the success of large language models (LLM) and their demonstration of scaling laws \citep{kaplan2020scalinglawsneurallanguage}. The premise is that training larger models on vast datasets enhances performance. This idea has been central to computer vision, where datasets like ImageNet \citep{5206848} have enabled pre-training in both supervised and unsupervised settings, leading to breakthroughs in model design and training. While pre-trained encoders extract valuable features from both natural and satellite images, key differences exist: Natural images use three bands (RGB), whereas satellite sensors capture multiple wavelengths, often generating images with tens or even hundreds of bands (hyperspectral data). Additionally, satellite image resolution varies significantly across sensors, complicating or even preventing direct model transfer.\\ \\
To address these challenges, we introduce PyViT-FUSE, a model designed to process multi-source, multi-modal images at their native resolutions. It enables flexible band selection through a channel fusion approach leveraging attention mechanisms. Unlike many EO foundation models that employ masking for self-supervised learning \citep{reed2023scalemaescaleawaremaskedautoencoder, astruc2024omnisatselfsupervisedmodalityfusion, jakubik2023foundationmodelsgeneralistgeospatial}, we adapt SwAV (Swapping Assignments between Views) \citep{caron2021unsupervisedlearningvisualfeatures}, eliminating the need for a decoder and pixel-space reconstruction. This avoids a key limitation of MAEs (Masked Autoencoders) \citep{he2021maskedautoencodersscalablevision}, which rely on accurate pixel-space reconstructions—a challenge in EO due to the less constrained nature of satellite imagery. As shown in Section \ref{sec:methodology}, SwAV also enables embeddings independent of input band combinations, allowing the same model to be applied across diverse downstream tasks. For instance, Sentinel-2 data may be sufficient for object detection but limited by cloud occlusion; integrating Sentinel-1 data mitigates this issue using the same model. We explore this important capability for a concrete segmentation task in appendix \ref{appx:application}.

\section{Methodology}
\label{sec:methodology}

In this section we describe our proposed model architecture (section \ref{sec:model_architecture}) and explain how we adapt the SwAV algorithm as self-supervised learning approach (section \ref{sec:model_training}). A core concept we leverage in this work is that multi-modal EO data is available co-registered for a given location, covering the same surface patch. We will refer to such a patch as area of view (AOV) with size $H \times W$ (in meters) which is the same for all considered modalities. Note that by modality we mean the data from a specific satellite source which can contain multiple bands, often at different resolutions.

\subsection{Model architecture}
\label{sec:model_architecture}

\begin{figure}[h]
	\centering
	\includegraphics[width=1.0\textwidth]{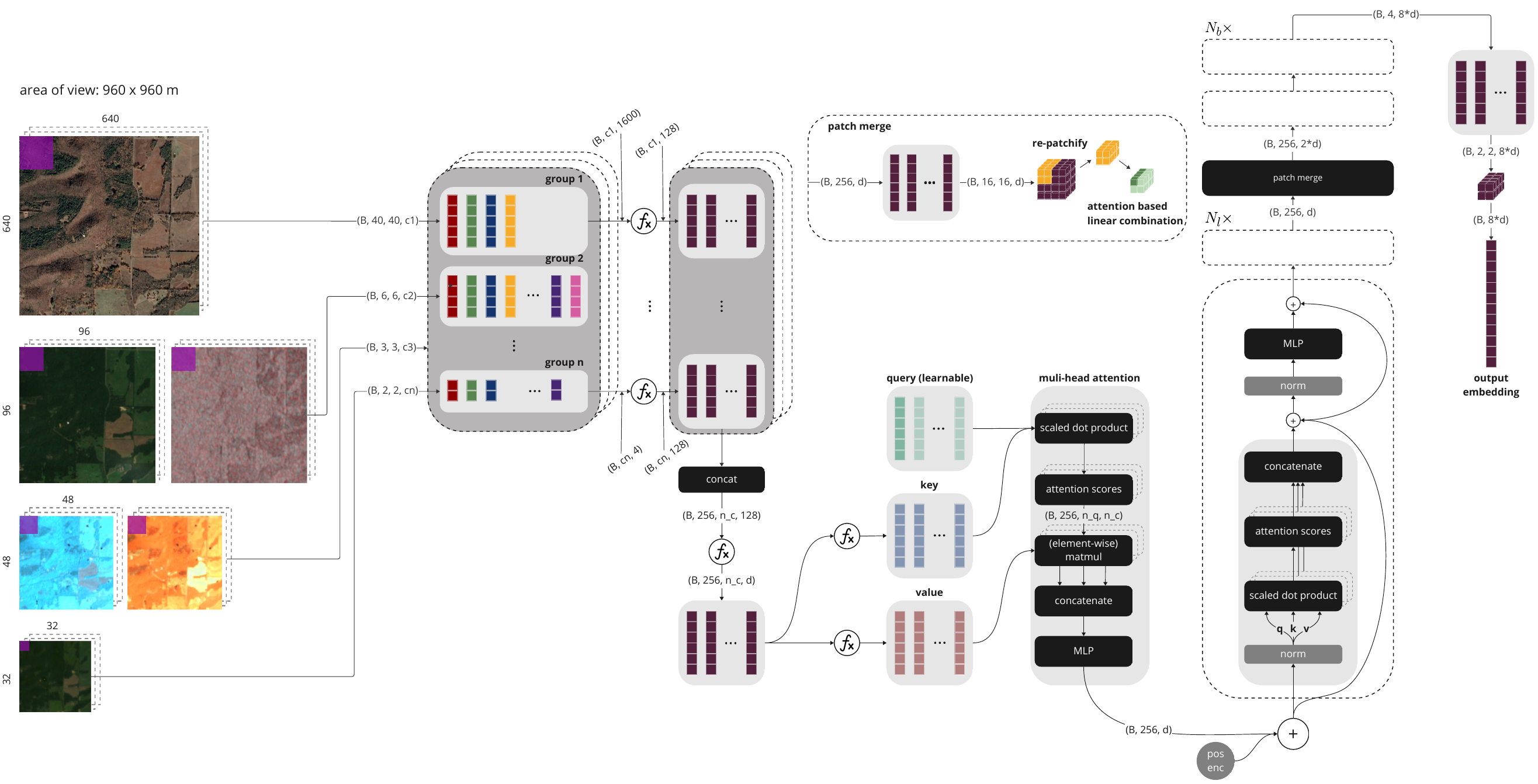}
	\caption{Model architecture of PyViT-FUSE consisting of three main components. An embedding representing the multi-modal input is generated independent of band combination.}
	\label{fig:model_architecture}
\end{figure}

The architecture of our model can roughly be divided into three main components: 1. Input module, 2. Fusion module, and 3. Pyramidal vision transformer as illustrated in Fig. \ref{fig:model_architecture}. Each of these components plays an important role in achieving the overall goal of generating embeddings enriched with extracted information from the available modalities.

\textbf{Input Module.} The input module processes each modality by dividing images into patches and transforming them into a shared feature space. Let $X_{i}^{r}$ be a 2D input array of size $H^{r} \times W^{r} = H/r \times W/r$ for band $i$ at resolution $r$. It is split into $n_{p} \times n_{p}$ patches, flattened into vectors of size $H^{r}/n_{p} \times W^{r}/n_{p}$, denoted as $x_{i}^{r}$. Grouping all $x_{i}^{r}$ from modality $M_{j}$ forms the sub-set $S^{M_{j}}=\{x_{0}^{r_{0}}, x_{1}^{r_{0}}, x_{2}^{r_{1}}, \dots, x_{n}^{r_{m}}\}$, while the complete input set is $S=\{S^{M_{0}}, S^{M_{1}}, \dots, S^{M_{k}}\}$, where $k$ is the number of modalities, each with $n_{j}$ bands, for a total of $n_{B}=\sum_{j=0}^{k}n_{j}$. A single modality may contain bands at different resolutions $r_{0}, r_{1}, \dots r_{m}$. Processing all bands at their native resolution without resampling is a key feature of our approach. Each raw pixel vector $x_{i}^{r}$ is then transformed into a common feature space via two learnable linear projections $\mathcal{P}_{i}^{1}$ and $\mathcal{P}_{i}^{2}$, yielding $\hat{x}_{i}=\mathcal{P}_{i}^{2}(\mathcal{P}_{i}^{1}(x_{i}^{r}))$ with feature dimension $d$. Using two projections instead of one allows setting a hidden layer dimension $d_{r}$ such that both layers have comparable parameter counts, ensuring sufficient capacity to learn transformations independently of band resolution.\\ \\
\textbf{Fusion Module.} The input module outputs a batch tensor of shape $(B, n_{p}^{2}, n_{B}, d)$, where $B$ is the batch size. To feed this into the vision transformer, which processes a sequence of patch tokens, the band dimension is fused into a single representation via attention. The fused representation is a weighted sum of input bands, $\hat{x}=\sum_{i}w_{i}\hat{x}_{i}$, where weights $w_{i}$ correspond to attention scores. A multi-head cross-attention layer performs this fusion: key and value vectors are derived from linear projections of input tokens, while the query is a learned vector. The query size is arbitrary but should be large enough to effectively learn band fusion. Each attention head uses the same query but learns individual attention scores ($w_{i}$). The outputs from all heads are concatenated and passed through a linear layer. As shown in Section \ref{sec:experiments_and_results}, this method enables visualization of band importance at each patch position, as attention weights determine how much information from each band propagates into subsequent model layers.\\ \\
\textbf{Transformer Module.} At this stage, the data is a tensor of shape $(B, n_{p}^{2}, d)$, processed through multiple multi-head self-attention layers. We follow the transformer encoder of \citet{vaswani2023attentionneed} but use pre-norm instead of post-norm, as in \citet{xiong2020layernormalizationtransformerarchitecture}. Additionally, we introduce a pyramidal structure where, after $N_{l}$ self-attention layers, spatial patch merging is applied, similar to CNN pooling \citep{NIPS2012_c399862d}. This uses the fusion module to merge patches via attention, increasing feature dimension by a factor of $s$. As illustrated in Fig. \ref{fig:model_architecture}, patches are first reassembled spatially, then re-patchified into $s \times s$ groups before fusion. Like band fusion, this results in a linear combination of neighboring feature vectors weighted by attention scores. Setting $s=2$ reduces the sequence length by a factor of four after each merge step. Repeating $N_{l}$ self-attention layers followed by patch merging $N_{b}$ times yields an output feature map with $n_{p}/2^{N_{b}-1}$ feature vectors of dimension $d \cdot 2^{N_{b}-1}$. This pyramidal structure compresses information into abstract representations, enabling downstream tasks that utilize feature maps at different depths.

\subsection{Model training}
\label{sec:model_training}
\begin{figure}[h]
	\centering
	\includegraphics[width=1.0\textwidth]{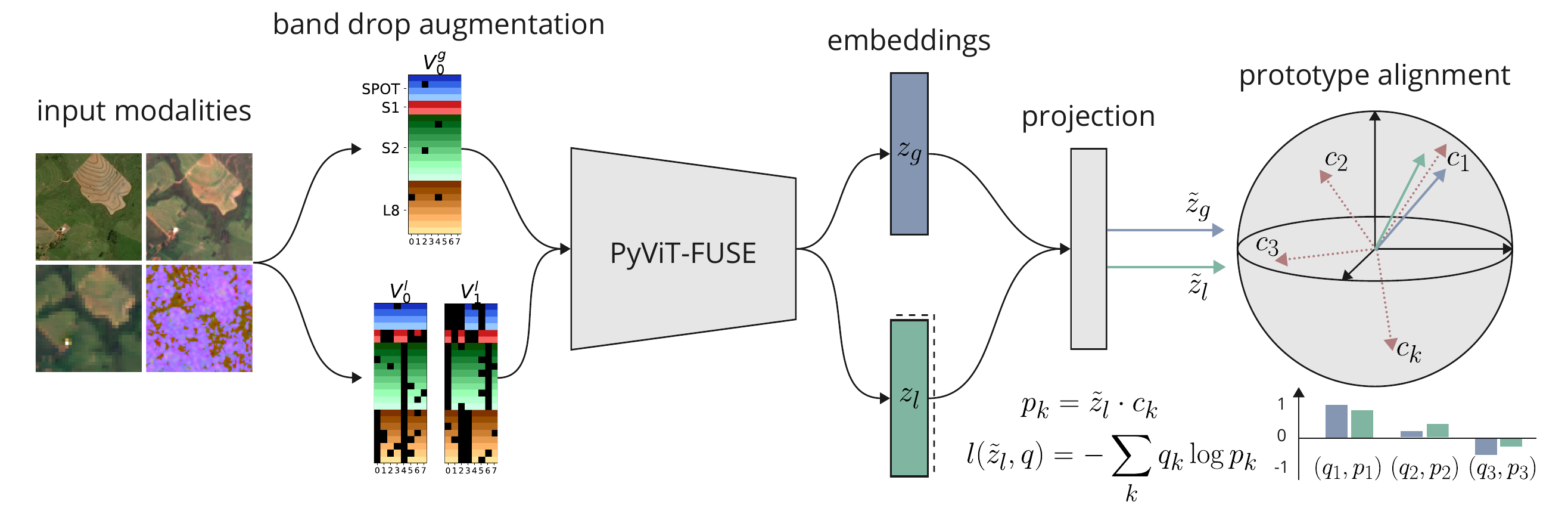}
	\caption{Illustration of band drop data augmentation and self-supervised training with SwAV algorithm. Band drop is visualized by coding the sensor as color hue and its bands as different shades for 8 batches. Blacked out parts indicate dropped channels.}
	\label{fig:swav}
\end{figure}

We adopt SwAV as a self-supervised training method, where the encoder generates similar embeddings for augmented views of the same input. Unlike contrastive methods \citep{chen2020simpleframeworkcontrastivelearning}, SwAV aligns embeddings to prototype vectors instead of relying on negative samples. Fig. \ref{fig:swav} illustrates SwAV, and we refer to \citet{caron2021unsupervisedlearningvisualfeatures} and appendix \ref{appx:model_training} for details. While the original paper uses multi-crop augmentation to generate local views, this is unsuitable for satellite images, where small crops may have distinct semantics. Instead, we exploit remote sensings unique multi-sensor observations, generating local views by randomly dropping modality sub-sets $S^{M_{j}}$ and individual bands $x_{i}^{r}$. Dropped bands are replaced with a learnable empty token, as shown in Fig. \ref{fig:swav}. This augmentation strategy enables SwAV to handle multi-sensor data, leading to a model that generalizes across different input band combinations. We carefully tune hyper-parameters such as the softmax temperature, learning rate, number of prototype vectors (set to 512 in this work) and batch size to avoid mode collapse.

\textbf{Dataset and training details.} For a proof of concept, we curated a globally sampled dataset of $\sim$1M samples (see appendix \ref{appx:dataset} for a map of the sample distribution). We choose an AOV of 960~m and include the following sensors/modalities: \textit{Airbus SPOT} (4 bands), \textit{Sentinel-1} (2 bands), \textit{Sentinel-2} (10 bands) and \textit{Landsat-8} (8 bands) for a total of 24 bands with resolutions ranging from 1.5 to 30~m/pixel. The models internal representation dimension is set to $d=128$, the query size of the fusion modules to 4096, and it consists of $N_{l}=8$ transformer layers per block and $N_{b}=4$ total blocks. All attention layers (self- and cross-attention) consist of 8 heads. This configuration results in a model with $\sim$103M parameters (3.8M for the input module, 1M for the fusion module and 98M for the ViT). We train the model on 8 A10G GPUs with a batch size of 256 using a stochastic gradient descent (SGD) optimizer with momentum of 0.9 for 30 epochs. The total training time was $\sim$1 week or $\sim$1344 GPU hours.

\section{Experiments and Results}
\label{sec:experiments_and_results}

\begin{figure}[h]
	\centering
	\includegraphics[width=1.0\textwidth]{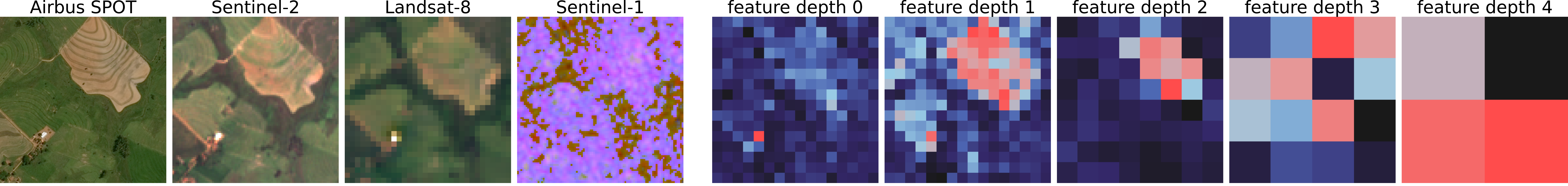}
	\caption{Sample input visualized as RGB image for each modality and corresponding averaged feature maps at the output of each ViT pyramid block. Colors indicate low (black) to high (red) activations.}
	\label{fig:feature_maps}
\end{figure}

We first inspect the feature maps generated by the trained model at the output of each ViT pyramid block. Fig. \ref{fig:feature_maps} shows an input sample where we display each modality as RGB image and the corresponding feature maps averaged across the feature dimension (see appendix \ref{appx:feature_maps_attention_scores} for more samples and detailed feature maps). It is clear that the model has learned to extract spatial features well.

\begin{figure}[h]
	\centering
	\includegraphics[width=0.8\textwidth]{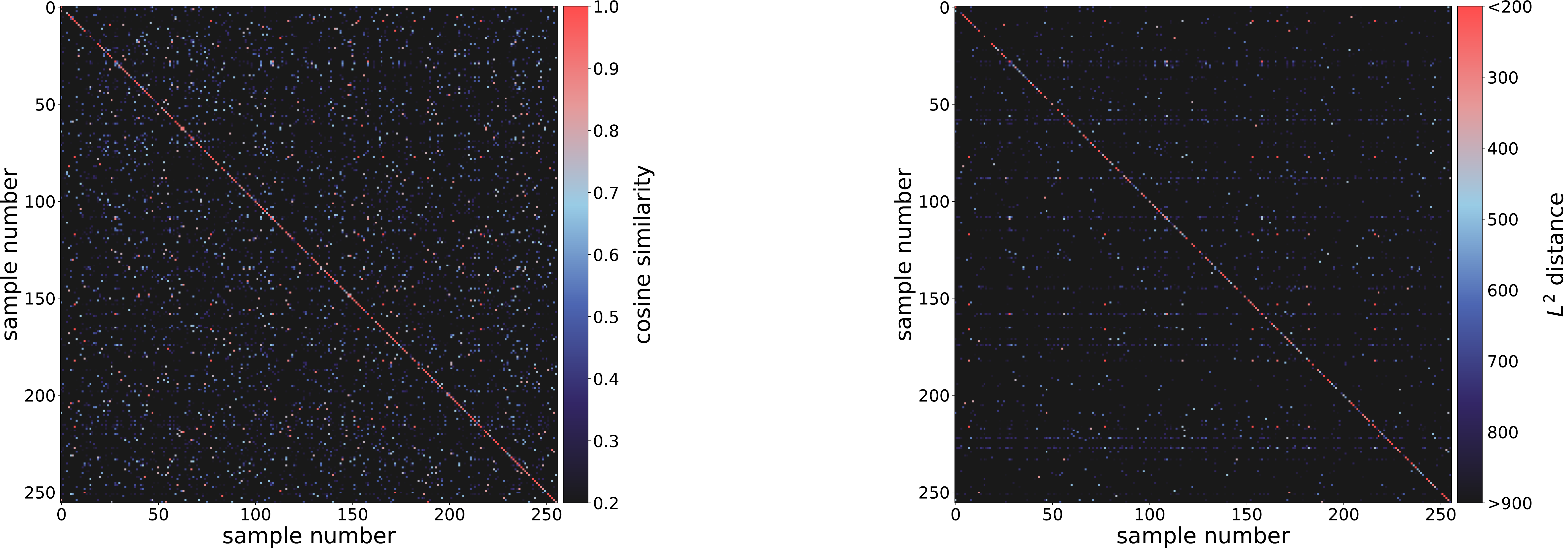}
	\caption{Cosine similarity (left) and $L^{2}$ distance (right) between the embeddings of the global and a local view for a sample against all other samples in the batch.}
	\label{fig:similarity_distance}
\end{figure}

A more quantitative assessment of the models learned skills is to inspect the regularity of the output embeddings. The objective of the learning task is for the model to generate embeddings that are close together in feature space if the inputs contain similar information as is the case for different band combinations of the same input sample. Given a global view of sample $i$ ($V_{i}^{g}$) and a local view of sample $j$ ($V_{j}^{l}$) within a batch, we can determine the cosine similarity $\sigma(V_{i}^{g}, V_{j}^{l})$ as well as $L^{2}$ distance $d_{2}(V_{i}^{g}, V_{j}^{l})$, where we expect $\sigma$ to be high and $d_{2}$ to be low if $i=j$ and opposite if $i \neq j$. Fig. \ref{fig:similarity_distance} illustrates both metrics as a matrix where $i,j$ correspond to the rows and columns. The images demonstrate that the model has learned to generate similar embeddings for views of the same surface patch with different band combinations which is the expected behavior.

\begin{figure}[h]
	\centering
	\includegraphics[width=1.0\textwidth]{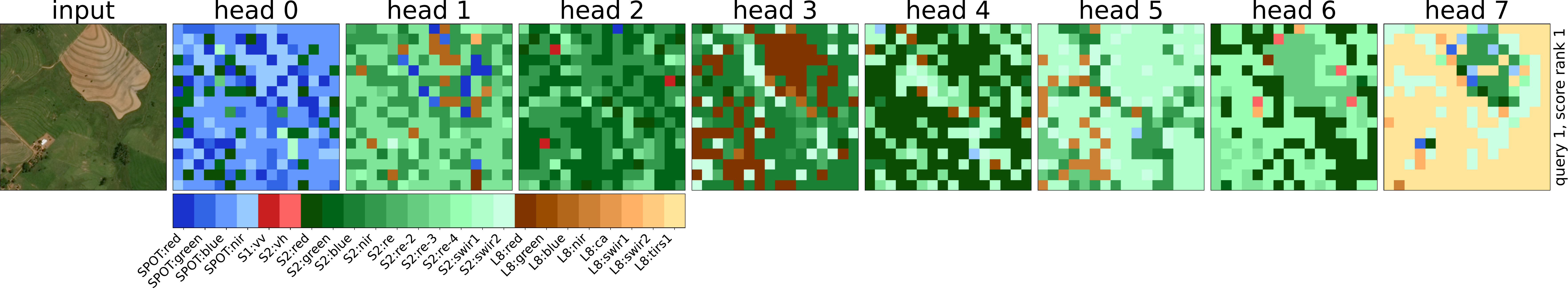}
	\caption{Visualization of attention scores for each head of the fusion module. The color corresponds to the band with the highest score indicating the corresponding importance of the band.}
	\label{fig:attention_scores}
\end{figure}

The fusion model allows the visualization and interpretation of the relative importance of the various bands at a given location. Fig. \ref{fig:attention_scores} show the attention scores of each head of the fusion module for the same sample as in Fig. \ref{fig:feature_maps}. We use a color coding that represents the sensor and band which has the highest attention score at the respective location, hence indicating the importance of the respective band. Visually, it is apparent that different heads learned to pay more attention to certain bands for specific features of the input data which is consistent across samples (see appendix \ref{appx:feature_maps_attention_scores}).

\section{Conclusion and Outlook}
\label{sec:conclusion_and_outlook}
We demonstrate a novel approach for the fusion of mixed-resolution, multi-modal satellite imagery, its interpretability due to attention mechanism and leverage an innovative concept for self-supervised training of a large model on vast amounts of data. As the next step, we will run a number of benchmark tests to quantify the model capability across different tasks as well as scale the methodology.

\bibliography{iclr2025_conference}

\begin{thebibliography}{13}
\providecommand{\natexlab}[1]{#1}
\providecommand{\url}[1]{\texttt{#1}}
\expandafter\ifx\csname urlstyle\endcsname\relax
  \providecommand{\doi}[1]{doi: #1}\else
  \providecommand{\doi}{doi: \begingroup \urlstyle{rm}\Url}\fi

\bibitem[Astruc et~al.(2024)Astruc, Gonthier, Mallet, and Landrieu]{astruc2024omnisatselfsupervisedmodalityfusion}
Guillaume Astruc, Nicolas Gonthier, Clement Mallet, and Loic Landrieu.
\newblock Omnisat: Self-supervised modality fusion for earth observation, 2024.
\newblock URL \url{https://arxiv.org/abs/2404.08351}.

\bibitem[Caron et~al.(2021)Caron, Misra, Mairal, Goyal, Bojanowski, and Joulin]{caron2021unsupervisedlearningvisualfeatures}
Mathilde Caron, Ishan Misra, Julien Mairal, Priya Goyal, Piotr Bojanowski, and Armand Joulin.
\newblock Unsupervised learning of visual features by contrasting cluster assignments, 2021.
\newblock URL \url{https://arxiv.org/abs/2006.09882}.

\bibitem[Chen et~al.(2020)Chen, Kornblith, Norouzi, and Hinton]{chen2020simpleframeworkcontrastivelearning}
Ting Chen, Simon Kornblith, Mohammad Norouzi, and Geoffrey Hinton.
\newblock A simple framework for contrastive learning of visual representations, 2020.
\newblock URL \url{https://arxiv.org/abs/2002.05709}.

\bibitem[Deng et~al.(2009)Deng, Dong, Socher, Li, Li, and Fei-Fei]{5206848}
Jia Deng, Wei Dong, Richard Socher, Li-Jia Li, Kai Li, and Li~Fei-Fei.
\newblock Imagenet: A large-scale hierarchical image database.
\newblock In \emph{2009 IEEE Conference on Computer Vision and Pattern Recognition}, pp.\  248--255, 2009.
\newblock \doi{10.1109/CVPR.2009.5206848}.

\bibitem[Fujita et~al.(2023)Fujita, Ancona, Kramer, Straka, Gautreau, Garrity, Robson, Diffendorfer, and Hoen]{solarpv2024}
K.S. Fujita, Z.H. Ancona, L.A. Kramer, M.~Straka, T.E. Gautreau, C.P. Garrity, D.~Robson, J.E. Diffendorfer, and B.~Hoen.
\newblock United states large-scale solar photovoltaic database (v2.0, august, 2024): U.s. geological survey and lawrence berkeley national laboratory data release, 2023.
\newblock URL \url{https://energy.usgs.gov/uspvdb/}.

\bibitem[He et~al.(2021)He, Chen, Xie, Li, Dollár, and Girshick]{he2021maskedautoencodersscalablevision}
Kaiming He, Xinlei Chen, Saining Xie, Yanghao Li, Piotr Dollár, and Ross Girshick.
\newblock Masked autoencoders are scalable vision learners, 2021.
\newblock URL \url{https://arxiv.org/abs/2111.06377}.

\bibitem[Jakubik et~al.(2023)Jakubik, Roy, Phillips, Fraccaro, Godwin, Zadrozny, Szwarcman, Gomes, Nyirjesy, Edwards, Kimura, Simumba, Chu, Mukkavilli, Lambhate, Das, Bangalore, Oliveira, Muszynski, Ankur, Ramasubramanian, Gurung, Khallaghi, Hanxi, Li, Cecil, Ahmadi, Kordi, Alemohammad, Maskey, Ganti, Weldemariam, and Ramachandran]{jakubik2023foundationmodelsgeneralistgeospatial}
Johannes Jakubik, Sujit Roy, C.~E. Phillips, Paolo Fraccaro, Denys Godwin, Bianca Zadrozny, Daniela Szwarcman, Carlos Gomes, Gabby Nyirjesy, Blair Edwards, Daiki Kimura, Naomi Simumba, Linsong Chu, S.~Karthik Mukkavilli, Devyani Lambhate, Kamal Das, Ranjini Bangalore, Dario Oliveira, Michal Muszynski, Kumar Ankur, Muthukumaran Ramasubramanian, Iksha Gurung, Sam Khallaghi, Hanxi, Li, Michael Cecil, Maryam Ahmadi, Fatemeh Kordi, Hamed Alemohammad, Manil Maskey, Raghu Ganti, Kommy Weldemariam, and Rahul Ramachandran.
\newblock Foundation models for generalist geospatial artificial intelligence, 2023.
\newblock URL \url{https://arxiv.org/abs/2310.18660}.

\bibitem[Kaplan et~al.(2020)Kaplan, McCandlish, Henighan, Brown, Chess, Child, Gray, Radford, Wu, and Amodei]{kaplan2020scalinglawsneurallanguage}
Jared Kaplan, Sam McCandlish, Tom Henighan, Tom~B. Brown, Benjamin Chess, Rewon Child, Scott Gray, Alec Radford, Jeffrey Wu, and Dario Amodei.
\newblock Scaling laws for neural language models, 2020.
\newblock URL \url{https://arxiv.org/abs/2001.08361}.

\bibitem[Kingma \& Ba(2017)Kingma and Ba]{kingma2017adammethodstochasticoptimization}
Diederik~P. Kingma and Jimmy Ba.
\newblock Adam: A method for stochastic optimization, 2017.
\newblock URL \url{https://arxiv.org/abs/1412.6980}.

\bibitem[Krizhevsky et~al.(2012)Krizhevsky, Sutskever, and Hinton]{NIPS2012_c399862d}
Alex Krizhevsky, Ilya Sutskever, and Geoffrey~E Hinton.
\newblock Imagenet classification with deep convolutional neural networks.
\newblock In F.~Pereira, C.J. Burges, L.~Bottou, and K.Q. Weinberger (eds.), \emph{Advances in Neural Information Processing Systems}, volume~25. Curran Associates, Inc., 2012.
\newblock URL \url{https://proceedings.neurips.cc/paper_files/paper/2012/file/c399862d3b9d6b76c8436e924a68c45b-Paper.pdf}.

\bibitem[Reed et~al.(2023)Reed, Gupta, Li, Brockman, Funk, Clipp, Keutzer, Candido, Uyttendaele, and Darrell]{reed2023scalemaescaleawaremaskedautoencoder}
Colorado~J. Reed, Ritwik Gupta, Shufan Li, Sarah Brockman, Christopher Funk, Brian Clipp, Kurt Keutzer, Salvatore Candido, Matt Uyttendaele, and Trevor Darrell.
\newblock Scale-mae: A scale-aware masked autoencoder for multiscale geospatial representation learning, 2023.
\newblock URL \url{https://arxiv.org/abs/2212.14532}.

\bibitem[Vaswani et~al.(2023)Vaswani, Shazeer, Parmar, Uszkoreit, Jones, Gomez, Kaiser, and Polosukhin]{vaswani2023attentionneed}
Ashish Vaswani, Noam Shazeer, Niki Parmar, Jakob Uszkoreit, Llion Jones, Aidan~N. Gomez, Lukasz Kaiser, and Illia Polosukhin.
\newblock Attention is all you need, 2023.
\newblock URL \url{https://arxiv.org/abs/1706.03762}.

\bibitem[Xiong et~al.(2020)Xiong, Yang, He, Zheng, Zheng, Xing, Zhang, Lan, Wang, and Liu]{xiong2020layernormalizationtransformerarchitecture}
Ruibin Xiong, Yunchang Yang, Di~He, Kai Zheng, Shuxin Zheng, Chen Xing, Huishuai Zhang, Yanyan Lan, Liwei Wang, and Tie-Yan Liu.
\newblock On layer normalization in the transformer architecture, 2020.
\newblock URL \url{https://arxiv.org/abs/2002.04745}.

\end{thebibliography}
\bibliographystyle{iclr2025_conference}

\newpage
\appendix
\section{Model and training details}
\label{appx:training_details}

\subsection{Dataset}
\label{appx:dataset}

\begin{figure}[h]
	\centering
	\includegraphics[width=1.0\textwidth]{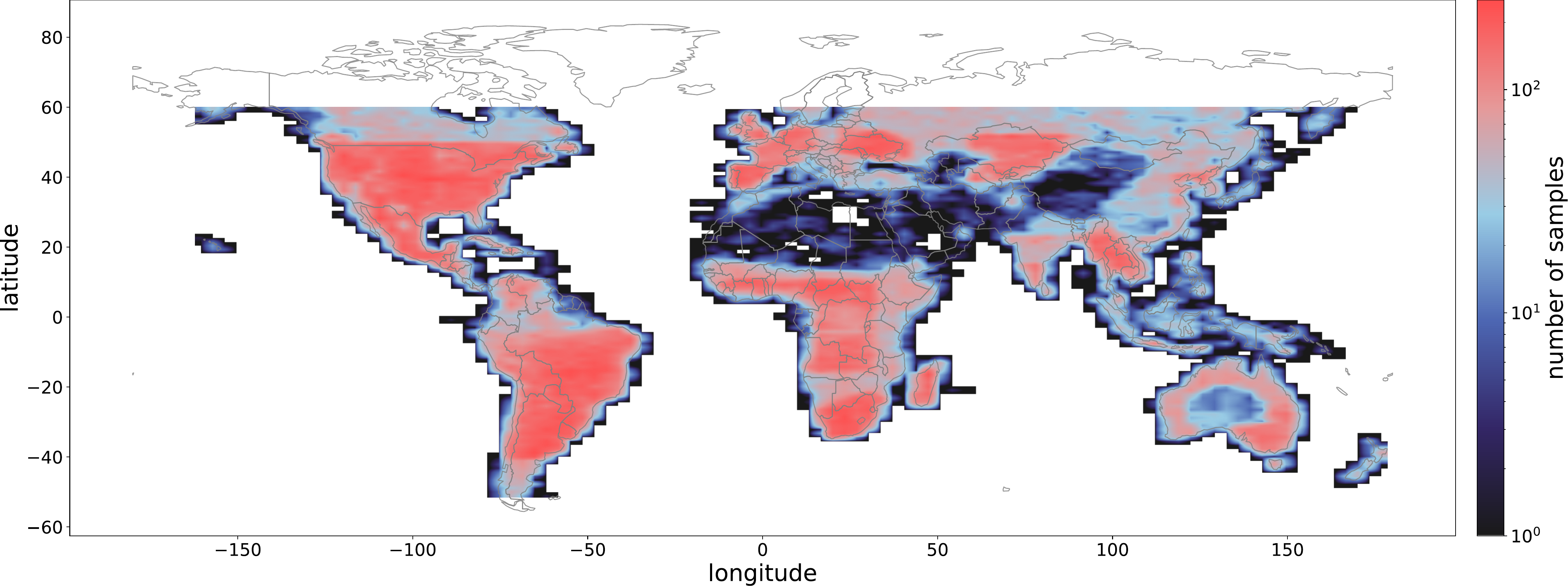}
	\caption{Distribution of samples gathered globally for an initial dataset.}
	\label{fig:dataset_distribution}
\end{figure}

For a proof of concept, we curated a dataset that includes four modalities (Airbus-SPOT, Sentinel-1, Sentinel-2 and Landsat-8) with a total of 24 bands. We gathered $\sim$1M samples (a total of 5.8~TB) at random locations around the global where all data sources are available. We only select samples where the cloud fraction is $<$7\% for all modalities. We remove samples in arid regions at a rate of 0.8 since they are largely cloud free which would lead to an oversampling of very similar image patches. Fig. \ref{fig:dataset_distribution} shows the distribution of samples.

\subsection{Model training}
\label{appx:model_training}

\begin{figure}[h]
	\centering
	\includegraphics[width=1.0\textwidth]{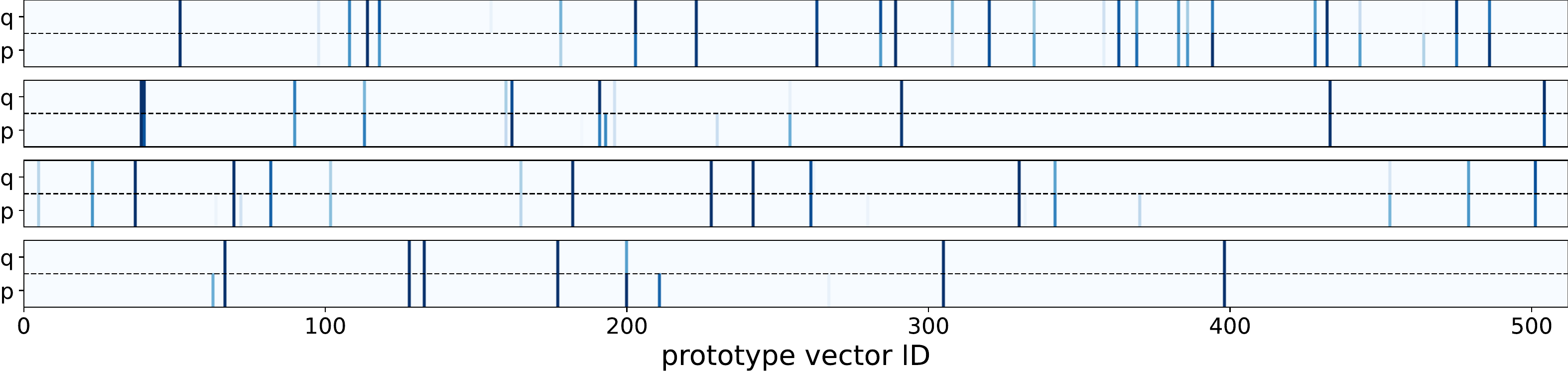}
	\caption{Visualization of alignment with prototype vectors for the global (q) and local (p) view of three samples.}
	\label{fig:prototype_alignment}
\end{figure}

Model training is performed using the self-supervised SwAV algorithm. We use a band drop augmentation and other image augmentations, such as additive and multiplicative noise, for generating global and local views. Global views should mostly contain uncorrupted inputs or weak augmentations, such as few single bands dropped, while the local views are created with heavy augmentations including dropping of entire input modalities. Training with SwAV requires $n_{g}$ global views and $n_{l}>n_{g}$ local views from which the loss is calculated as
\begin{equation}
    L=\sum_{i=0}^{n_{g}}\sum_{j=0}^{n_{l}}(1-\delta_{ij})l(\tilde{z}_{j},q_{i})
\end{equation}
There are two mode collapses in which the model learns a trivial solution: 1. Always predict the same cluster assignment or 2. assign clusters uniformly across all prototype vectors, independent of the input data. The first case is addressed by using the Sinkhorn-Knopp algorithm to re-distribute the distribution of $q$. This method works best if the number of prototype vectors is $\ll B$ (batch size). This requirement invokes some compute limitations as the number of prototype vectors should be relatively high (thousands) which also requires the batch size to be high. This limitation can be somewhat reduced by introducing a queue that stores the projected embeddings from previous batches. Current prototype assignments are then re-computed for the current training step using the queue while the queue is updated with the current batch at the end of the step. In this work we choose a batch size of 256 and a queue size of 4 allowing the number of prototype vectors to be 512.\\ \\
After the model is trained, we can verify its ability to assign the same prototypes to both the global and local view. Fig. \ref{fig:prototype_alignment} provides a visualization of the prototype assignments for three different samples. The top part of each strip corresponds to the codes provided by the global view while the bottom part are the corresponding predictions based on the local views. The strength of alignment against each prototype is indicated by the color of the line. This demonstrates that different samples are aligned with different prototypes, only a few prototypes have strong alignments and $p$ and $q$ are aligned with the same prototypes which is the expected behavior.

\section{Additional experimental results}
\label{appx:results}

\subsection{Feature maps and attention scores}
\label{appx:feature_maps_attention_scores}

\begin{figure}[h]
	\centering
	\includegraphics[width=1.0\textwidth]{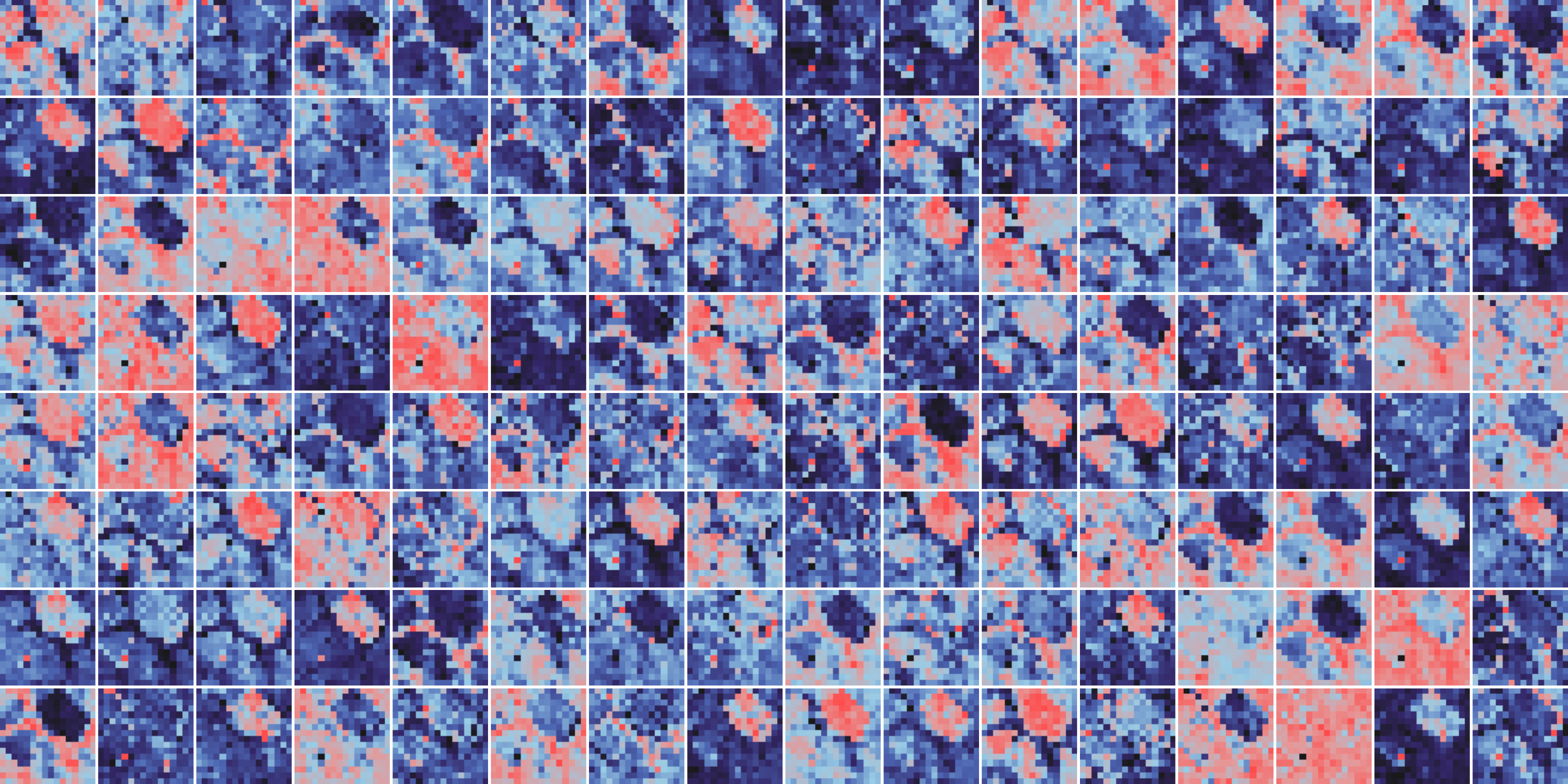}
	\caption{Map of all features from the output of the first ViT block, color coded from low (black) to high (red) activation.}
	\label{fig:all_feature_maps}
\end{figure}

Fig. \ref{fig:feature_maps} shows the averaged feature maps at the output of each ViT block. This, however, does not illustrate all the learned features for a given output. In Fig. \ref{fig:all_feature_maps} we show all 128 feature maps at the output of the first ViT block illustrating the various spatial structures the model picks up on. Deeper feature maps are less interpretable since spatial components are further compressed.

\begin{figure}[h]
	\centering
	\includegraphics[width=1.0\textwidth]{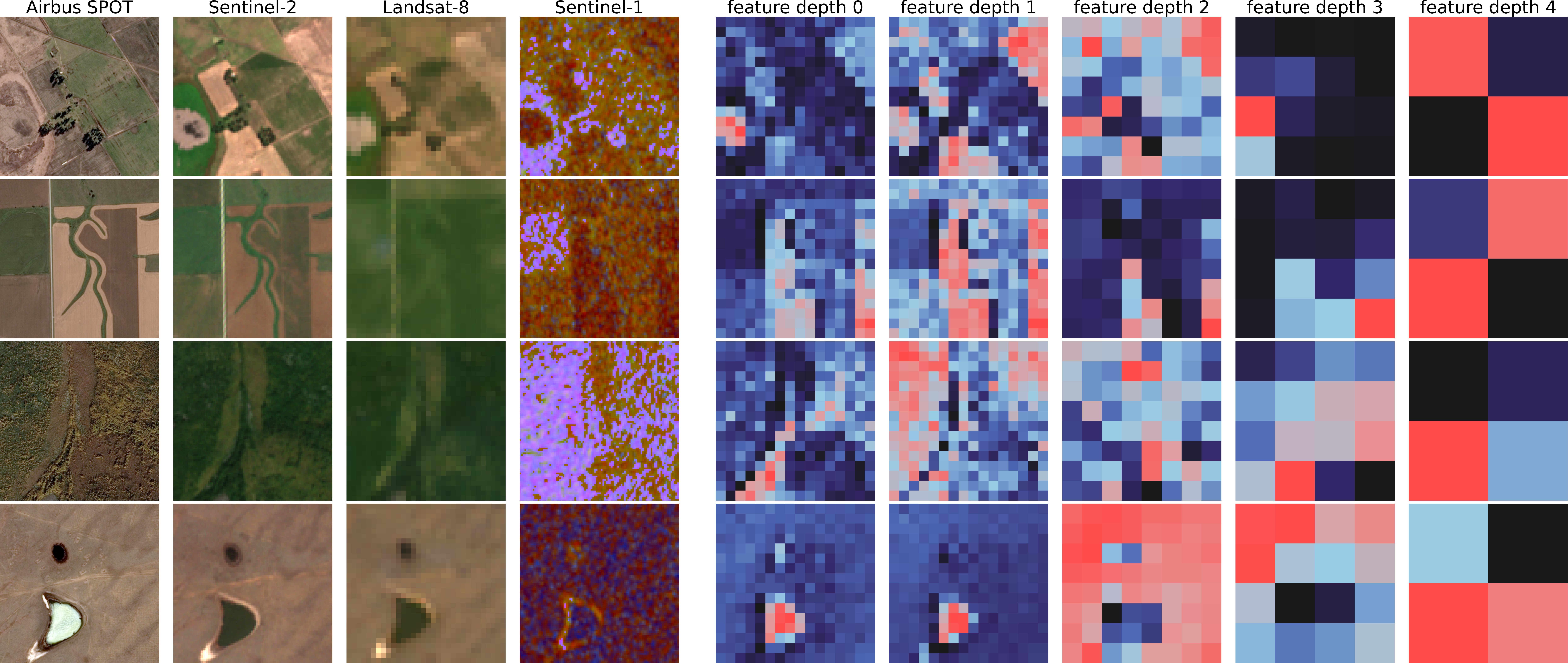}
	\caption{Visualization of the input as RGB image of four samples and corresponding averaged feature maps at the output of each ViT block.}
	\label{fig:samples_feature_maps_all}
\end{figure}

\begin{figure}[h]
	\centering
	\includegraphics[width=1.0\textwidth]{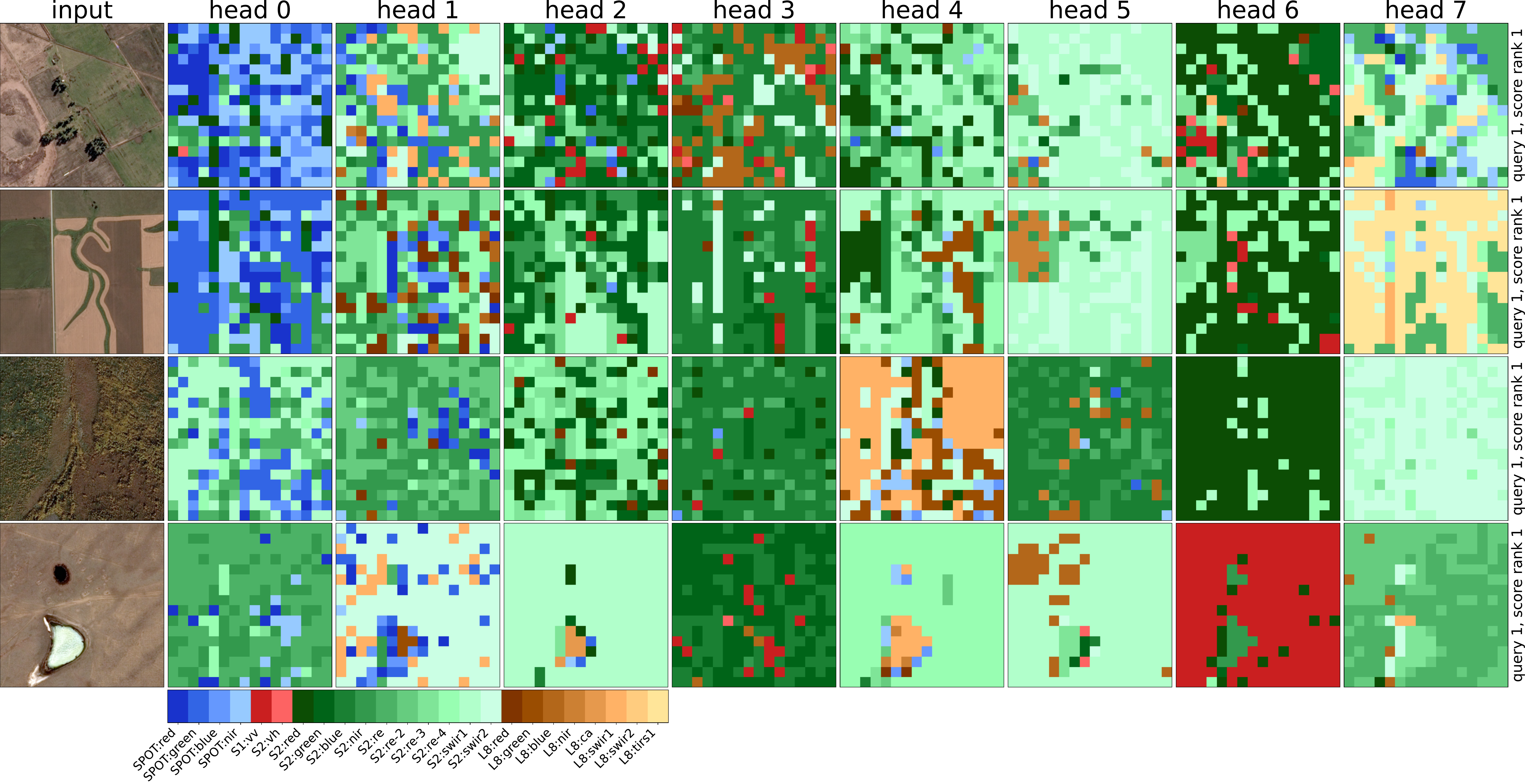}
	\caption{Visualization of the attention scores from all heads of the fusion module for four samples. The color represents the sensor and band with the highest attention score at the given position.}
	\label{fig:attention_scores_all}
\end{figure}

We show additional input samples and corresponding averaged feature maps in Fig. \ref{fig:samples_feature_maps_all}, similar to Fig. \ref{fig:feature_maps}. For the same samples we illustrate the band importance given by the highest attention scores in each head of the fusion module in Fig. \ref{fig:attention_scores_all}. It is apparent that different heads tend to focus on specific modalities and bands as well as spatial features.

\subsection{Application}
\label{appx:application}

\begin{figure}[h]
	\centering
	\includegraphics[width=1.0\textwidth]{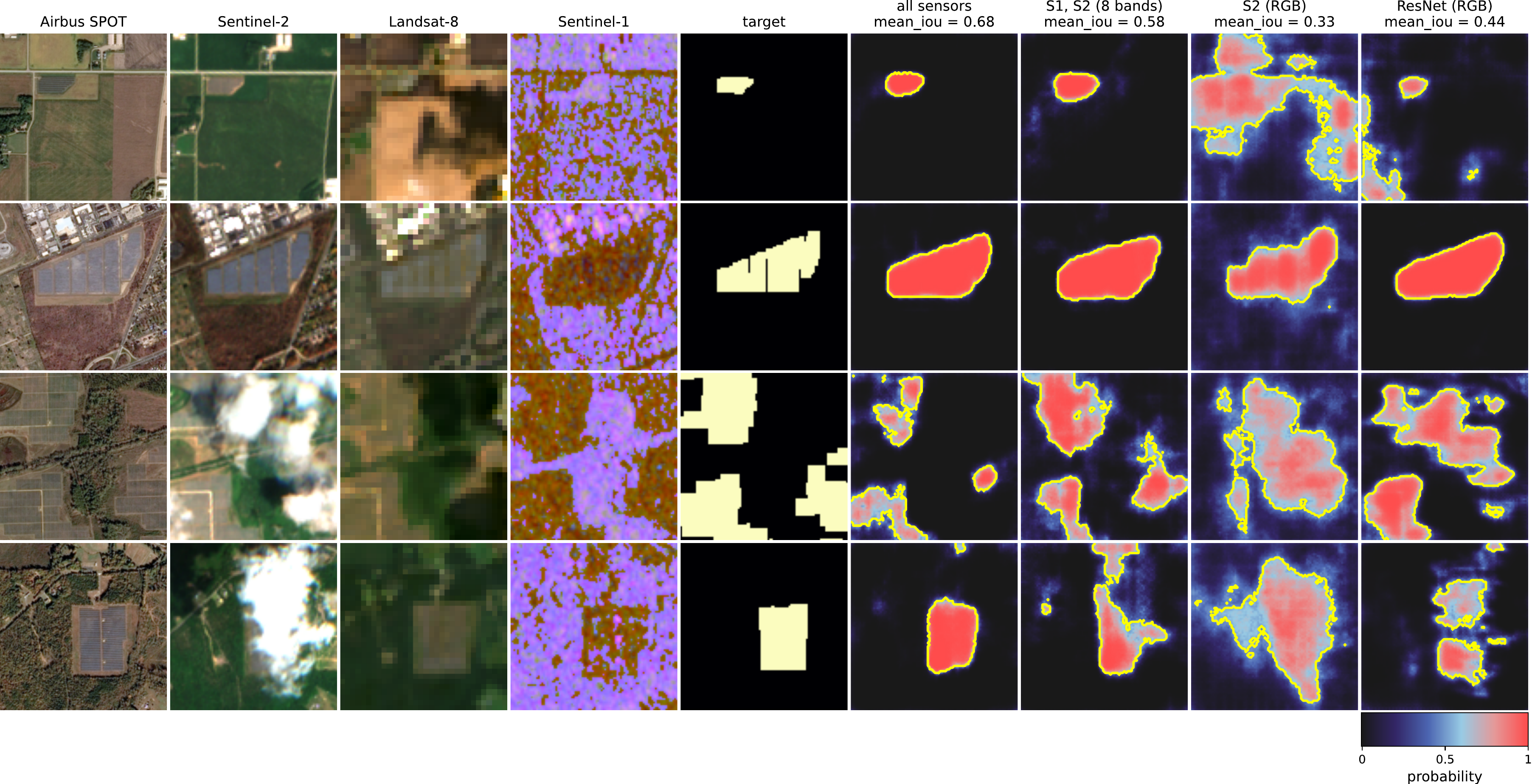}
	\caption{Input of all modalities visualized as RGB image (first four columns from the left), corresponding target map and the output probability maps of the fine-tuned PyViT-FUSE model on the solar PV dataset for three different input band combinations. The right column corresponds to the output probability map of a fine-tuned FPN with a pre-trained ResNet-50 encoder as a comparison.}
	\label{fig:solarpv_outputs}
\end{figure}

\begin{figure}[h]
	\centering
	\includegraphics[width=1.0\textwidth]{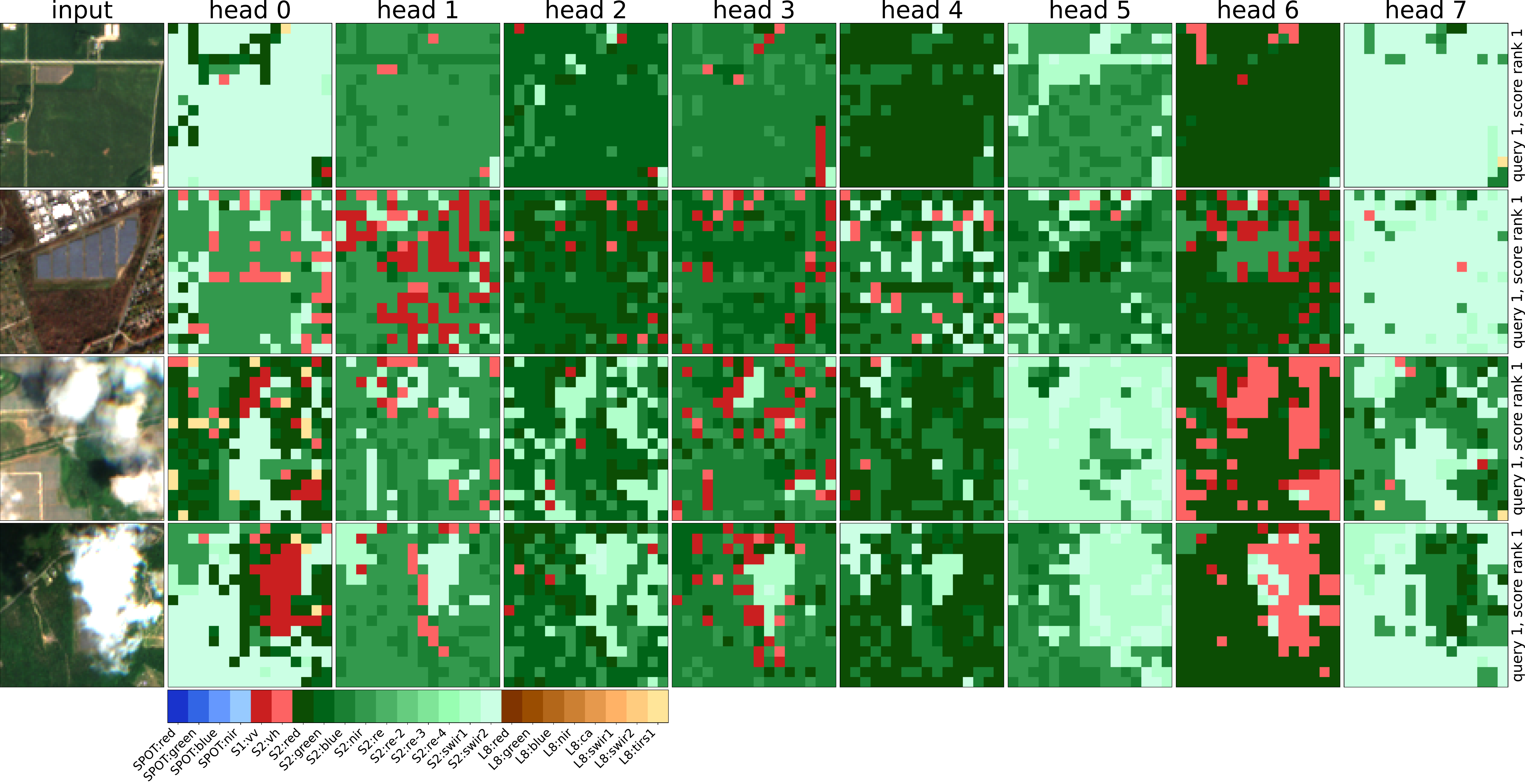}
	\caption{Visualization of the attention scores of the fusion module where the color corresponds to the band with the highest score. The input consists of 6 bands from Sentinel-2 and 2 bands from Sentinel-1.}
	\label{fig:solarpv_attention_scores}
\end{figure}

\begin{figure}[h]
	\centering
	\includegraphics[width=1.0\textwidth]{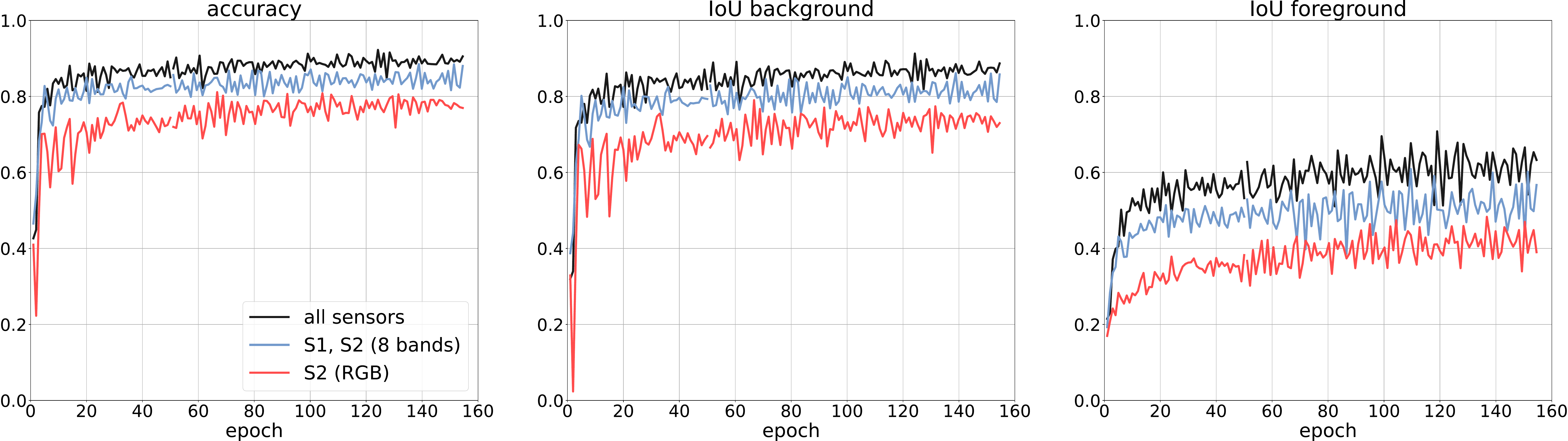}
	\caption{Training plots of accuracy, background IoU and foreground IoU from fine-tuning PyViT-FUSE on the solar PV dataset for three different input band combinations.}
	\label{fig:solarpv_training_plots}
\end{figure}

We demonstrate a concrete application of our foundation model where we use a dataset containing the segmentation masks of large solar photovoltaic installations \citep{solarpv2024}. In total, the dataset contains 4185 facilities aggregated into 1597 samples across the United States, including 142 samples with no positive pixels, for which we gather the imagery from all modalities after but close to the facilities installation date. We randomly select 100 of the samples and set it aside for validation. We attach a simple decoder network which processes the encoder feature maps from all depths by a series of convolutional + up-sample + concatenate layers until the desired output resolution is reached. This is commonly referred to as Feature Pyramid Network (FPN). As output resolution, we choose 10~m/pixel resulting in an output map of size 96 $\times$ 96 pixels. We train the network by minimizing the binary-crossentropy loss using the Adam optimizer \citep{kingma2017adammethodstochasticoptimization} in two steps: First, all weights are frozen except the decoder, resulting in 3.7M trainable parameters. After 50 epochs, we un-freeze the weights in the fusion module in order to adapt for this specific task. The second stage consists of 4.7M trainable parameters and continues for another 100 epochs. During training, the learning rate is decreased from an initial value of 1e$^{-4}$ to 1e$^{-7}$ according to a cosine function.\\ \\
We perform the same training procedure for different input band combinations in order to study the effect of variable information content for solving the specific task. The first fine-tuned model uses all modalities as input while the second model only uses 6 bands (red, green, blue, NIR, SWIR1, SWIR2) of Sentinel-2 and 2 bands (VV, VH) of Sentinel-1. The third model further reduces the amount of input information and only uses 3 bands (red, green, blue) of Sentinel-2. We also train a CNN based FPN with a ResNet-50 encoder as backbone pre-trained on ImageNet as baseline. Since the pre-trained model expects three input bands, we use red, green and blue of Sentinel-2. Fig. \ref{fig:solarpv_outputs} illustrates the output probability maps of all models for four specific samples. They include various sizes of targets. The first two samples are considered easy as the input is very clean while the last two samples are hard due to cloud obstruction in the Sentinel-2 imagery. We also show the contour lines (yellow) of the binary output map at a threshold of 0.5. There are two main observations:
\begin{enumerate}
    \item The detection accuracy increases as more bands are added to the input. This is quantified as the foreground IoU at an output threshold of 0.5 which increases from 0.33 for S2 (RGB) to 0.58 for S1, S2 (8 bands) and 0.68 for all sensors. The baseline model achieves an IoU of 0.44. This is further illustrated in Fig. \ref{fig:solarpv_training_plots} where we plot the training curves for the metrics accuracy, background IoU and foreground IoU during training of the various models.
    \item For the inputs which contain obstructions in Sentinel-2 imagery, the model accuracy is generally worse. Both the baseline model and our model, using only Sentinel-2 RGB as input, perform poorly on these hard samples. However, the models which utilize bands from other modalities perform much better, highlighting the importance and benefits of fusing different modalities.
\end{enumerate}
The visualization of the attention scores of the fusion module is particularly interesting for the model using Sentinel-1 and Sentinel-2 as shown in Fig \ref{fig:solarpv_attention_scores}. It shows that the bands with highest attention scores belong to the modalities used as input while the empty tokens of all other bands do not carry any valuable information. Moreover, it shows that specific heads pick up on the locations of obstruction by assigning higher attention scores to the Sentinel-1 bands, notable in head 6 for the last two samples where the bands colored in red align with the clouds.

\end{document}